\documentclass{article}
\usepackage[utf8]{inputenc}
\usepackage[T1]{fontenc}
\usepackage{times,amsmath,amssymb,amsthm,graphicx,booktabs,tabularx,array}
\usepackage{microtype,xcolor,tikz}
\usetikzlibrary{arrows.meta,positioning,calc,fit}
\usepackage[hidelinks]{hyperref}
\hypersetup{pdftitle={AgentEval: Evidence-Bounded Assessment of Stateful Agents},pdfauthor={Ruiqi Gao},pdfsubject={Stateful agent assessment, evidence qualification and governed evaluation workflows}}
\usepackage{iclr2027_conference}
\iclrfinalcopy
\definecolor{ink}{HTML}{20384B}
\definecolor{teal}{HTML}{256E73}
\definecolor{pale}{HTML}{EFF4F5}
\newcommand{\AES}{\textsc{AgentEval}}

\newcommand{\unk}{\mathord{?}}

\newcolumntype{Y}{>{\raggedright\arraybackslash}X}

\newcommand{\pageend}{\par}
\expandafter\def\csname AESResult@AESReferenceDisagreement\endcsname{0}
\expandafter\def\csname AESResult@B4Difference\endcsname{0}
\expandafter\def\csname AESResult@FullPolicyUnknown\endcsname{192}
\expandafter\def\csname AESResult@HoldoutCompleted\endcsname{192}
\expandafter\def\csname AESResult@MaskEvaluable\endcsname{192}
\expandafter\def\csname AESResult@MaskGatedUnknownInvalid\endcsname{192}
\expandafter\def\csname AESResult@MaskRawFailPartial\endcsname{192}
\expandafter\def\csname AESResult@MaskRawUnknownInvalid\endcsname{0}
\expandafter\def\csname AESResult@ObjectiveReferenceDecidable\endcsname{192}
\expandafter\def\csname AESResult@ObjectiveReferenceUnknown\endcsname{0}
\expandafter\def\csname AESResult@PairedComplete\endcsname{96}
\expandafter\def\csname AESResult@PairedDiscordant\endcsname{5}
\expandafter\def\csname AESResult@deepseek-v4.1-flash|C0|Allocated\endcsname{32}
\expandafter\def\csname AESResult@deepseek-v4.1-flash|C0|ObjectiveSuccess\endcsname{31}
\expandafter\def\csname AESResult@deepseek-v4.1-flash|C1_typeI_paraphrase|Allocated\endcsname{32}
\expandafter\def\csname AESResult@deepseek-v4.1-flash|C1_typeI_paraphrase|ObjectiveSuccess\endcsname{32}
\expandafter\def\csname AESResult@kimi-k3|C0|Allocated\endcsname{32}
\expandafter\def\csname AESResult@kimi-k3|C0|ObjectiveSuccess\endcsname{32}
\expandafter\def\csname AESResult@kimi-k3|C1_typeI_paraphrase|Allocated\endcsname{32}
\expandafter\def\csname AESResult@kimi-k3|C1_typeI_paraphrase|ObjectiveSuccess\endcsname{30}
\expandafter\def\csname AESResult@qwen3.8-flash|C0|Allocated\endcsname{32}
\expandafter\def\csname AESResult@qwen3.8-flash|C0|ObjectiveSuccess\endcsname{31}
\expandafter\def\csname AESResult@qwen3.8-flash|C1_typeI_paraphrase|Allocated\endcsname{32}
\expandafter\def\csname AESResult@qwen3.8-flash|C1_typeI_paraphrase|ObjectiveSuccess\endcsname{31}
\expandafter\def\csname AESResult@CoreTargetRequests\endcsname{2626}
\expandafter\def\csname AESResult@CoreTargetInputTokens\endcsname{14043657}
\expandafter\def\csname AESResult@CoreTargetOutputTokens\endcsname{428996}
\expandafter\def\csname AESResult@CoreSimulatorRequests\endcsname{1823}
\expandafter\def\csname AESResult@CoreSimulatorInputTokens\endcsname{878831}
\expandafter\def\csname AESResult@CoreSimulatorOutputTokens\endcsname{736159}
\expandafter\def\csname AESResult@CoreExecutionWallSeconds\endcsname{6409.12}
\expandafter\def\csname AESResult@CoreTaskRetries\endcsname{0}

\newcommand{\RealResult}[1]{%
  \ifcsname AESResult@#1\endcsname
    \csname AESResult@#1\endcsname
  \else
    \PackageError{AgentEval-results}{Missing accepted measured result: #1}%
      {Generate the result definitions from the accepted full audit; do not supply invented defaults.}%
  \fi
}

\title{AgentEval: Evidence-Bounded Assessment\\of Stateful Agents}
\author{Ruiqi Gao\\Alibaba Cloud, Alibaba Group}
\begin{document}
\maketitle
\lhead{\small Preprint \hfill AgentEval: Evidence-Bounded Assessment}
\begin{abstract}
Task completion alone does not determine whether a stateful agent satisfied its obligations, and a recorded judgment does not guarantee that its supporting evidence remains valid. We present \AES{}, a system that carries a contract-bound assessment record through tool execution, state and event capture, report qualification, and case maintenance. It separates what the agent did, what the evidence establishes, and whether the conclusion is supported for present use. On 32 tool-world administrations, independent checks agree on all 30 complete episodes, distinguishing satisfied, violated, and unresolved obligations; two interrupted administrations are retained. Controlled interventions on 48 evidence views expose 36 binding faults missed by hash-only checks without changing agent behavior. Scenario Studio connects reviewed cases to 12 two-turn reports with 24 independently checked answers. A separate public-benchmark study on 192 held-out executions reveals ten completed return subgoals within broader task failures and isolates the effect of missing evidence. These studies show how an executable assessment record makes scope, evidential support, and report reuse inspectable across an agent evaluation workflow.
\end{abstract}

\section{Introduction}
\label{sec:intro}
A refund agent reports success. Did money move exactly once? Did it modify only the authorized order? If ownership was ambiguous, did it obtain a trusted answer before acting? And can the assessment still be used after its world or evidence changes? Each question concerns the meaning of the result. Stateful agents make this distinction consequential: the same final answer can accompany different persistent effects, and a plausible completion message cannot establish those effects.

Interactive benchmarks make agent behavior measurable through executable tasks and environments \citep{tau,agentbench,webarena,swe}. Policy-aware, procedure-aware, and reliability-oriented evaluations already go beyond task completion \citep{dojo,stweb,pae,reliability}. Recent work asks whether retained evidence supports a score and whether assessment remains valid as agents and benchmarks change \citep{evidencebounds2026,trustlayer2026,frozenjudges,benchmarkmaintenance2026}. The practical challenge is to carry these distinctions through task construction, real execution, evidence capture, assessment, report reuse, and subsequent case maintenance.

We frame this challenge as \emph{evidence-bounded assessment}. An execution has a fixed history, while its admissibility for present use depends on the evidence and contract that still support it. The key design choice is to carry one bound assessment record across execution, report inspection and case maintenance: each stage depends on the same declared obligations and supporting artifacts. An assessment answers three linked questions: \emph{What happened? Did it satisfy the declared obligations? What conclusion is supported now?} A declared objective names the quantity being assessed; execution conditions specify authorization, interaction, faults, and budgets. Evidence supplies the state and events needed to decide. Version and provenance bindings delimit where a retained conclusion applies.

\AES{} implements this view through a trusted host for permission, transaction and clarification worlds, evaluator-only evidence capture, and content-addressed assessments. Report inspection rechecks support while preserving history. The web workbench connects these paths to source-bound case maintenance and Scenario Studio: business intent, reviewed revisions, multi-turn execution and failure-informed planning.

Our contributions are \textbf{(1) a contract-bound assessment record and trusted execution host}, linking goal and process obligations to tool actions and evaluator evidence; \textbf{(2) report qualification that preserves historical judgments}, exposing when changes in evidence or measurement bindings prevent present use; and \textbf{(3) provenance-bound case maintenance integrated with an operator workbench}, connecting source eligibility, reviewed revisions, and versioned use. Tool-world executions and evidence interventions test assessment and qualification; reviewed multi-turn cases test the operator workflow. A separate public-benchmark study examines objective scope and evidence loss.

\begin{figure}[t]
\centering
\includegraphics[width=\linewidth]{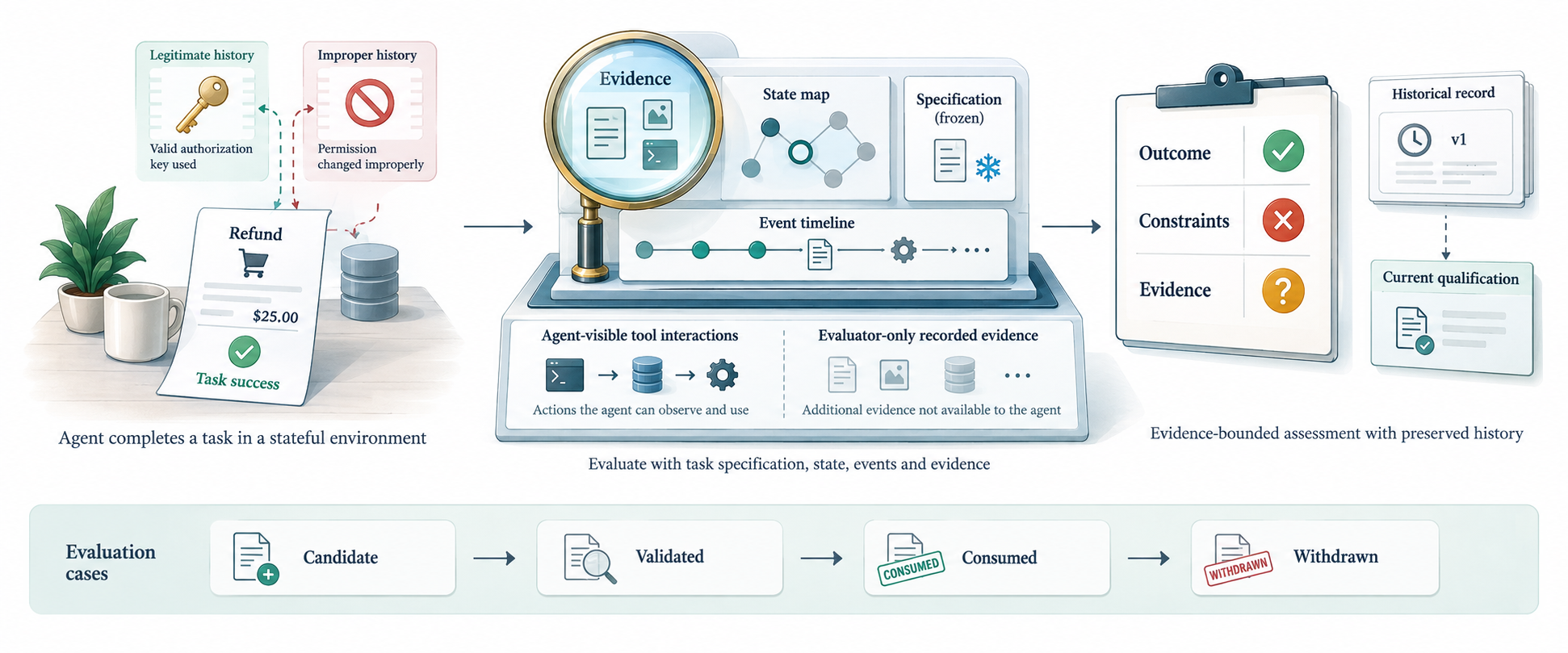}
\caption{\textbf{A completion becomes an assessment through evidence.} This illustrative workflow checks a refund against persistent effects and declared obligations. Bound state, event and version records support the judgment and its current qualification. Governed cases connect reviewed admission, actual consumption and withdrawal while retaining historical evidence.}
\label{fig:teaser}
\end{figure}

\section{Related work}
\label{sec:related}
\paragraph{Interactive tasks and operational evaluation.}
AgentBench, WebArena, SWE-bench and $\tau$-bench supply executable tasks \citep{agentbench,webarena,swe,tau}. AgentDojo, ST-WebAgentBench and procedure-aware evaluation assess utility, attacks, policy and interaction constraints \citep{dojo,stweb,pae}. Reliability and behavioral testing motivate repeated, condition-specific measurement \citep{reliability,science,checklist,toolrobust}. \AES{} retains distinct full-task and declared-subgoal assessments.

\paragraph{Evidence and evaluator validity.}
Judge calibration and trajectory evaluation examine the assessor itself \citep{judge,reward}; active testing addresses evaluation efficiency \citep{active}. Evidence-supported bounds add a reporting layer to existing benchmarks \citep{evidencebounds2026}, while the Trust Layer audits saved computations, completion claims and repeat stability \citep{trustlayer2026}. Claim-dependent reliability and transported judge calibration further delimit score interpretation \citep{claimreliability2026,frozenjudges}. \AES{} binds the assessment contract before execution and carries its identity through current report inspection and versioned case admission. Our contribution is this executable connection; three-valued labels and finite bounds are established tools within it.

\paragraph{Evaluation workflows and maintained suites.}
An earlier AgentEval uses LLM agents to generate and verify task-utility criteria \citep{agentevalutility2024}. AgentSuite audits benchmark components; Inspect and Promptfoo provide evaluation workflows \citep{agentsuite2026,inspect_evalsets,promptfoo_testcases}; practical evaluation and generated tests emphasize harnesses and failure traces \citep{anthropic_evals2026,bloom2025}. Benchmark maintenance addresses evolving coverage \citep{benchmarkmaintenance2026}. \AES{} connects stateful execution, evidence requalification and provenance-bound admission, measuring these paths separately from task-generation gains.

\pageend
\section{Assessment objects and semantics}
\label{sec:setup}
An agent configuration $A$ comprises its model, prompts, orchestration, memory, tools, and stopping rules. A condition $c=(w,\mathcal O,u,q,f,b)$ specifies a world, observation interface, authorization, trusted requester context, declared faults, and execution budget. Execution yields an interaction trajectory $\tau=(o_0,a_0,r_0,\ldots,o_T)$ and a full episode $\xi=(\tau,s_0,s_T,\mathcal E)$ containing initial/final world state and events. Evaluator evidence $e$ is a retained view of $\xi$; some facts are privileged and never enter the target observation interface.

\paragraph{Declared obligations.}
A task contract $\Gamma$ names the goal and applicable permission, side-effect, transaction, and interaction requirements. When necessary context defines every obligation as a binary predicate, its nonempty obligation set $J$ gives
\begin{equation}
Y^*_{\Gamma}(\xi)=\bigwedge_{j\in J}g_j(\xi),\qquad
Y^*_{G}(\xi)=\bigwedge_{j\in J_G}g_j(\xi),\quad \varnothing\ne J_G\subseteq J.
\label{eq:success}
\end{equation}
The subset relation requires the same goal definition and fully specified context. In the transaction world, the full refund contract additionally checks amount, settlement, declared faults and world invariants. Clarification has a conditional, partially defined contract: a bound trusted reply determines the document obligation; a request without that reply can remain unresolved. Its unconditional final-document goal is a different estimand, not necessarily a subset of this partial contract. The historical retail study checks three defined return obligations within requests that can contain other actions. Each contract names its own scope.

\paragraph{Assessment record.}
We represent a retained assessment as
\begin{equation}
\mathcal R=(A,\Gamma,c,v,e,\widehat Y,m,h),
\label{eq:context}
\end{equation}
where $v$ binds target, task, world and checker identities; $\widehat Y$ contains check results; $m$ records measurement status; and $h$ records provenance and content digests. The record retains what was judged at execution and whether its support currently qualifies for use.

\paragraph{Three separate questions.}
For a fully defined contract, behavioral truth $Y^*_{\Gamma}$ concerns what the agent did; evidence adequacy concerns whether $e$ determines it. A partial contract can additionally lack context needed to define an obligation. Current qualification $Q(\mathcal R,v_{\mathrm{now}})$ concerns whether retained bindings support the present report. Evidence loss cannot change past behavior; invalid bindings cannot be repaired by relabeling a historical PASS.

\pageend
\section{Evidence-bounded assessment in AgentEval}
\label{sec:method}
\subsection{From observations to executable checks}
A versioned specification declares the facts and obligations required by each check. The canonical verification registry retains \texttt{PASS}, \texttt{FAIL}, \texttt{PARTIAL}, \texttt{UNKNOWN}, and \texttt{INVALID}, with check-level reasons. The current tool-world host additionally uses trusted state and event predicates. We preserve native records rather than impose one undocumented status conversion on every entry point.

For a well-bound, fully defined conjunction, a three-valued assessment is
\begin{equation}
\widehat Y_{\Gamma}(e)=
\begin{cases}
+,&\text{every required obligation is established},\\
-,&\text{a required violation is established},\\
\unk,&\text{available facts establish neither conclusion}.
\end{cases}
\label{eq:verdict}
\end{equation}
Identity or contract mismatch is separately invalid; execution interruption is a measurement failure. An established violation need not disappear when an unrelated fact is missing. Conditional checks additionally preserve UNKNOWN when necessary requester context is absent; this does not certify that an uncreated document satisfied a binary goal. Current contracts appear in Appendix~\ref{app:current}; the historical adapter's distinct composition is retained in Appendix~\ref{app:algorithm}.

\paragraph{Trusted execution and evidence isolation.}
The host freezes task and measurement identities before running a registered target. The agent receives its task, projected observations, and tool schemas. World actions enforce permissions and transactions. A trusted responder binds clarification replies to a request token and task identity. Initial and final state, action events, model requests, and receipts form evaluator evidence. A content-addressed store binds these artifacts to run, subject, and specification digests. Expected answers, privileged state, and the independent checker remain outside the target request envelope.

\subsection{Current qualification without rewriting history}
Report inspection reopens retained evidence, specification and declared measurement identity. Abstracting the implemented checks,
\begin{equation}
Q(\mathcal R,v_{\mathrm{now}})=
I_{\mathrm{binding}}\wedge I_{\mathrm{integrity}}\wedge
I_{\mathrm{contract}}\wedge I_{\mathrm{dependency}}.
\label{eq:valid}
\end{equation}
Qualification requires matching run, subject and specification bindings, intact evidence bytes, a supported contract and the declared measurement dependency identity. It permits a current report, including unresolved outcomes, without resolving every obligation. Recorded execution and diagnoses remain unchanged; historical readback depends on artifact availability. Changes outside the declared measurement closure, such as presentation changes, need not invalidate an assessment. The checks do not establish arbitrary program or provider equivalence.

\begin{table}[t]
\centering\small
\caption{\textbf{Different records permit different uses.} Current qualification is checked independently of behavioral outcome.}
\label{tab:report-actions}
\begin{tabularx}{\linewidth}{@{}llY@{}}
\toprule
Behavior & Current support & Permitted interpretation\\
\midrule
PASS & Qualified & Report established contract satisfaction.\\
FAIL & Qualified & Report a violation; separately check case-source admission.\\
UNKNOWN & Qualified & Retain unresolved obligations and their declared reasons.\\
Any recorded status & Refused & Retain the historical record; withhold current report use.\\
\bottomrule
\end{tabularx}
\end{table}

\paragraph{Unresolved-status envelope.}
For $N=n_++n_-+n_{\unk}>0$ completed, independently admissible records, assigning unresolved statuses to zero or one gives
\begin{equation}
\left[\frac{n_+}{N},\frac{n_++n_{\unk}}{N}\right].
\label{eq:bounds}
\end{equation}
This is a label-accounting envelope. If latent binary outcomes are defined and established labels are sound, it is also a finite identification bound on satisfaction. Missing contract context alone does not establish those premises. It is neither a confidence interval nor a new estimator. Invalid bindings and interrupted administrations retain separate counts and a declared denominator policy.

\subsection{Governed case maintenance}
A versioned asset $z=(\Gamma,c,\mathrm{source},v_z)$ has an explicit lifecycle state. Source failure supports a proposal, not automatic correctness. Calibration uses family-specific positive, defect, and, where applicable, missingness controls. Admission records the approved revision and conditions. New consumption checks approval before registering a real run. Withdrawal prevents further use while retaining earlier evidence and consumption receipts.

Current worlds expose closed recipe families with bounded permission, amount, and document-content parameters. Parameter variation extends coverage within those families; it does not establish new task semantics or independent generalization. Historical judgments retain their original contracts throughout transitions. Case maintenance concerns evaluation assets; the evaluated agent remains a separate target.

\pageend
\subsection{Condition-indexed reports}
\label{sec:scope}
For condition $c$, a report names the evaluated objective, opportunity set, task components, versions, and supporting evidence. Decided-only satisfaction and coverage are
\begin{equation}
\widehat p_{\mathrm{dec}}=\frac{n_+}{n_++n_-},\qquad
\kappa=\frac{n_++n_-}{N}.
\label{eq:coverage}
\end{equation}
The first quantity is undefined without decisions and depends on which opportunities remain decidable. Under a partial contract it is a decided-label ratio, rather than an unconditional binary success rate. Reports retain unresolved reasons and qualification diagnoses alongside definite outcomes.

Comparisons preserve the intended relation between conditions. An equivalent wording transformation should retain goals and obligations; a permission or fault change can alter legitimate behavior. An evidence intervention freezes behavior and changes only the assessor's view. The historical study exercises wording and evidence contrasts; the current worlds expose controlled authorization, transaction-fault, and trusted-reply conditions. Repetitions, parameter variants, and evidence copies remain distinct from independent task structures.

A software claim ceiling further constrains report interpretation using target access, observability, and evidence. Internal causal explanations require more than outcome records, and comparisons require compatible evaluation envelopes. These mechanisms make scope explicit without replacing the assumptions needed for scientific inference.

\pageend
\section{System and operator workflow}
\label{sec:system}
\begin{figure}[t]
\centering
\includegraphics[width=\linewidth]{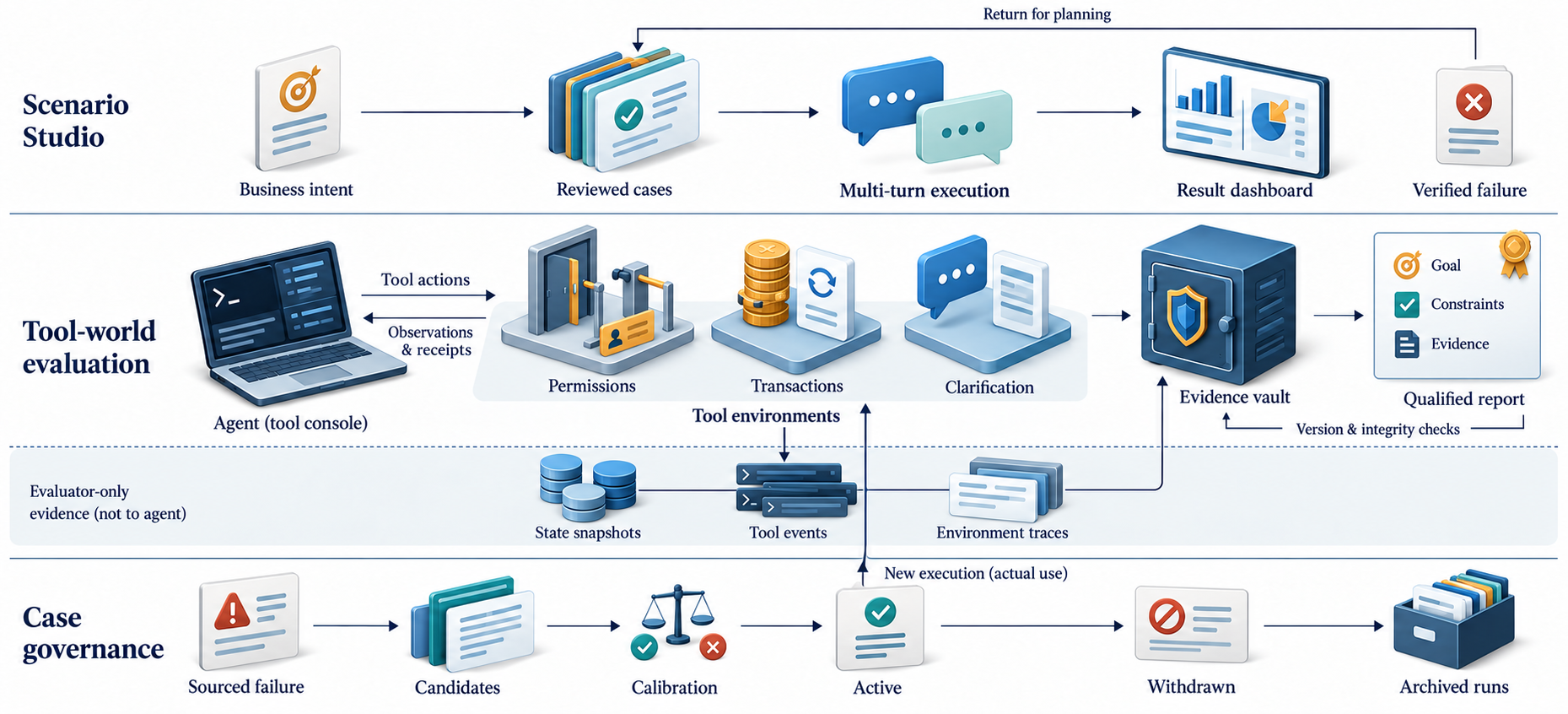}
\caption{\textbf{Three connected product workflows.} Scenario Studio constructs reviewed multi-turn cases. Tool-world evaluation separates agent-visible observations from privileged state and event evidence, then qualifies the retained report. Case governance admits and consumes assets in new runs; withdrawal preserves prior records. The paths share a registered execution host and evidence infrastructure, with distinct assessment contracts.}
\label{fig:architecture}
\end{figure}

\AES{} organizes evaluation into three connected workflows: reviewed multi-turn cases, stateful tool execution, and governed case maintenance (Figure~\ref{fig:architecture}). They share an execution host and evidence infrastructure while retaining their own task contracts. Appendix~\ref{app:current} details the interfaces and obligations. The public-benchmark study uses a separate retail adapter; its protocol and results are reported independently.

\paragraph{Real tool execution.}
The execution host supports Bailian model decisions through world tool actions and Qoder through a constrained native MCP session. Both connect to permission, transaction and clarification environments. Target registration binds the configuration before execution; admission checks, request budgets, deadlines, cancellation and recovery govern each run. The host captures actions and evidence and produces a durable report. Qoder and HTTP decision scaffolds constitute different agent configurations, so comparisons must account for their orchestration as well as the underlying model.

\paragraph{Evidence and reports.}
Initial and final state, events, model requests and tool receipts are stored as content-addressed evidence. Report inspection rechecks the declared bindings without reexecuting the target. Qualified evidence supports a current report; a refused binding leaves the historical judgment unchanged, subject to the availability of its artifacts. Suite comparisons additionally require compatible conditions and qualified paired reports.

\paragraph{Cases and governance.}
World recipes create bounded variants of permission, transaction and clarification tasks. A failure-derived candidate carries its source record, controls, scope and admission decision. Approved revisions can be used in new runs; withdrawal blocks further use while preserving prior evidence. The lifecycle maintains provenance and eligibility as cases change. Recipe parameters vary coverage within an existing task family.

\paragraph{Scenario Studio and the web view.}
An operator selects a project, describes business intent and rules, reviews proposed dimensions and resource recommendations, and imports or generates candidate cases. Each edit creates an immutable revision. Bailian execution assigns an independent session to each sample and preserves multi-turn history; the Studio Qoder path supports single-turn cases. Reports expose answers, per-turn and per-dimension results, and evidence qualification. Verified failures can inform a new plan through their source records. The chat grader uses exact-answer comparison after trimming whitespace. Consequently, generated cases require an answer check or operator review; stored rubrics are descriptive rather than an open-ended semantic judge.

The workbench also exposes project and target management, saved sets, controlled worlds, repeat runs, governed suites, diagnostic reports, and historical readback.

\pageend
\section{Experiments}
\label{sec:design}
\subsection{Questions and assessment units}
We examine three questions under fixed execution and assessment protocols. \textbf{RQ1: Contract semantics.} What do goal-only and full-contract checks establish from the same executions? \textbf{RQ2: Report qualification.} Which evidence and binding faults prevent present report use while behavior stays fixed? \textbf{RQ3: Operator workflows.} How does source eligibility govern case admission, and do reviewed multi-turn cases produce evidence-bound reports? A separate retail study examines objective scope in a public benchmark.

\paragraph{Real tool-world executions.}
E1 allocates 24 administrations: three families (permission, transaction, clarification), two forms (canonical and bounded recipe), two Bailian configurations and two repetitions. Both configurations use the original decision scaffold, tools and observation interface. Requested identifiers are \texttt{qwen3.8-flash} and \texttt{deepseek-v4.1-flash}; receipts retain returned identities and termination fields. Each episode permits eight requests, 512 output tokens per request and 180 seconds, with automatic retries disabled. Intrinsic step budgets and actual first-action fault exposure remain part of the contract. E2 adds eight matched clarification administrations changing trusted-reply availability. The executions use seed 7; parameter variants and repetitions remain within the declared task families. Full request settings and all interrupted outcomes are retained in the artifact.

\paragraph{Independent checks and comparators.}
A separately implemented checker reads task specifications, initial and final facts, events, provider receipts and externally bound identities. It evaluates the selected world invariants and task obligations without reading product verdicts. Goal-only and hash-only checks provide feature ablations. Positive, violation, missingness and forged-binding controls validate checker semantics before interpreting model outcomes. Reports distinguish unresolved behavior, invalid bindings and interrupted execution.

\paragraph{Evidence and binding interventions.}
E3 uses six predetermined E1 anchors: the first Qwen repetition of each form, regardless of outcome. Thirty host views alter raw availability, byte integrity, current instrument binding, registration binding or snapshot scope; eighteen direct views alter specification, seed or nested run binding. Isolated copies are inspected intact, edited and restored to the original semantic row values and evidence bytes. Current qualification, historical readability and independent eligibility are separate measures; the 48 views are not new agent executions.

\paragraph{Case lifecycle and Scenario Studio.}
The conditional E4 protocol selects the earliest eligible failure per family, then previews, proposes, validates, activates, consumes and withdraws a supported nonduplicate recipe. Without an eligible source, consumption remains unallocated. Studio instead uses six authored two-turn store-language programs, independently derived preexecution answers and immutable reviewed revisions. Both configurations run them, allocating 24 target answers. Planning, execution, software controls and offline checks retain separate units.

Detailed task declarations, controls, fault edits and lifecycle prerequisites accompany the frozen study manifest. The historical retail setup, its independent three-obligation return reference and its custom training-source split appear in Appendices~\ref{app:algorithm}--\ref{app:allocation}.

\pageend
\label{sec:scope-results}
\subsection{Actual execution, persistent effects and assessment}
All 32 allocated tool-world administrations were attempted. Thirty completed; two ended with an idempotency conflict when an agent reused a tool-budget key with different request parameters. All 93 actual provider requests returned and settled. The original errors remain in the allocation, without replacement trials.

\begin{table}[t]
\centering\small
\caption{\textbf{Actual current-world outcomes.} P/F/U count independently eligible, complete episodes only. Goal-only P/F uses the same episodes; interrupted allocations remain separate. Native and independent full-contract labels agree on all 30 complete episodes.}
\label{tab:current-world}
\begin{tabularx}{\linewidth}{@{}Yrrrr@{}}
\toprule
Condition / family & Allocated & Complete & Full P/F/U & Goal P/F\\
\midrule
E1: permission & 8 & 8 & 8/0/0 & 8/0\\
E1: transaction & 8 & 8 & 8/0/0 & 8/0\\
E1: clarification & 8 & 8 & 7/1/0 & 7/1\\
E2: absent trusted reply & 8 & 6 & 0/1/5 & 0/6\\
\midrule
Total & 32 & 30 & 23/2/5 & 23/7\\
\bottomrule
\end{tabularx}
\end{table}

The system and independently implemented full-contract checks agree on 23 PASS, two FAIL and five UNKNOWN records among the 30 complete episodes (Table~\ref{tab:current-world}). A correction to the independent checker's handling of refused actions was applied uniformly to all records, without changing any execution (Appendix~\ref{app:current-provenance}). Goal-only checking yields the same 23 passes but classifies all remaining complete episodes as failures. The distinction in this collection concerns unresolved contractual context, rather than goal attainment with a demonstrated process violation.

\paragraph{A requested action is not a committed effect.}
The first permission trace exhibiting authorization recovery has caller \texttt{alice}, but its first resolution request names \texttt{audit} as resolver. The world refuses the request, leaving identical pre/post state digests. A subsequent request names \texttt{alice}, emits the authorized event and resolves the ticket. The original assessment is PASS. Retaining proposals and actual effects explains this outcome: a forbidden request was blocked before it could violate the persistent-state contract. Final-state scoring omits the recovery; counting every proposal as an executed write misdescribes it.

\subsection{A trusted reply changes what can be established}
Of eight reply-availability pairs, six complete and two retain an interrupted absent-reply member. Five complete pairs change from PASS to UNKNOWN; one remains FAIL. In the five unresolved episodes, the agent requests clarification but receives no bound trusted answer. The unconditional document-goal ablation marks all six absent-reply episodes FAIL. The partial contract instead distinguishes missing context from failure to request clarification. UNKNOWN neither establishes document creation nor overrides binary goal failure. Target stochasticity prevents attributing the contrast solely to reply availability.

Of the 30 complete records, 25 have definite labels. Equation~\ref{eq:bounds} gives an unresolved-status envelope of $[23/30,28/30]=[76.7\%,93.3\%]$ and a decided-label ratio of $23/25$. The envelope accounts for five unresolved statuses. Because the absent-reply contract is partially defined, these endpoints do not identify a binary satisfaction rate for the same trajectories. The two interrupted administrations remain outside this conditional denominator.

\subsection{A retained judgment and its present use can diverge}
\begin{figure}[t]
\centering
\includegraphics[width=\linewidth]{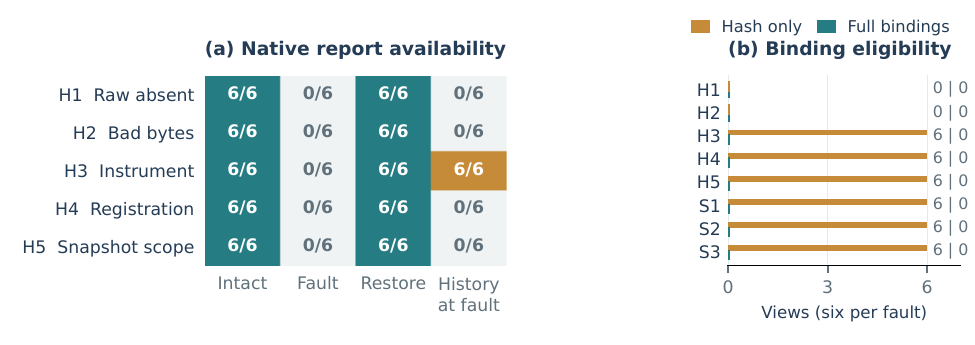}
\caption{\textbf{Present qualification depends on more than intact bytes.} (a) Original host report availability and historical readability on five faults. (b) Hash-only and full-binding eligibility on all eight faults; right-edge labels list their respective counts. Each fault has six fixed anchors; counts are edited evidence views, not agent executions. H1--H5 and S1--S3 are specified in Appendix~\ref{app:actual-interventions}.}
\label{fig:current-qualification}
\end{figure}

The 48 registered-run interventions comprise 30 host views and 18 direct-verifier views, each observed intact, edited and restored. All intact views are independently eligible; all edited views lose full binding eligibility. The host withdraws current report availability in all 30 edited views. Under the six instrument-binding edits, historical judgments remain readable while current reports are unavailable (Figure~\ref{fig:current-qualification}a). This is the distinction between preserving what was recorded and permitting its present use.

Hash-only checks accept 36 of the 48 edited views: their bytes remain consistent while the specification, seed, registration, instrument or scope no longer matches the assessment (Figure~\ref{fig:current-qualification}b). Full checks instead distinguish six unavailable views from 42 invalid bindings. Restoring each view recovers its original semantic evidence and report outputs. The direct verifier treats edited views as unavailable rather than behavioral failures. Report restoration is also distinct from qualification for cross-agent comparison, which remains unsupported under this study's registered-runtime interface.

\subsection{Qualified sources constrain case maintenance}
Source admission examines the two full-contract failures. Both lack the explicit requester reply required by the clarification recipe, so neither qualifies as a source. The admission rule therefore prevents unsupported case creation. Activation, subsequent use and withdrawal are implemented and covered by software controls, but this allocation provides no eligible source on which to measure their forward effect.

\subsection{A reviewed case becomes a real multi-turn report}
Scenario Studio completed an intent-planning call and executed the same reviewed revision with both configurations. All 12 two-turn samples produced reports, and all 24 target answers matched the program answers derived before execution. The actual first answer entered each second-turn history; sessions remained isolated across samples. Four primary dimensions partition the six programs: isolation and missing-value handling each have two cases, while overwrite and readback have one each. Figure~\ref{fig:studio-actual} illustrates the reviewed-case and conversation interfaces. These results validate execution and report binding for the defined program set, with operator-declared answers checked separately by the research interpreter.

\pageend
\label{sec:validity}
\subsection{Historical public-benchmark study: why objective scope matters}
The separately frozen retail study comprises 96 calibration and 192 held-out executions on eight and 16 actor/order components from the $\tau$-bench training split. Each component has three target configurations, two instruction conditions and two repetitions. The declared return subgoal checks order status, item set and refund method; native grading assesses the broader task.

Independent subgoal implementations agree on 187 satisfactions and five violations in all 192 holdout executions, with no unresolved references. Native grading reports 177 successes and 15 failures: ten achieved return subgoals coexist with full-task failure. One retained trace completes a thermostat return while omitting a separate mouse-order modification. This explains a scope distinction without treating either contract as universally preferable.

The instruction contrast retains 96 complete pairs and five discordances. Qwen has 31/32 successes in each condition, although two paired outcomes reverse in opposite directions. Condition-indexed records expose this difference hidden by equal aggregates. Target and simulated-user stochasticity prevent attributing it solely to wording.

With required goal fields withheld in 192 evaluator-facing copies, the raw checker assigns FAIL to every copy, while the completeness-aware adapter assigns UNKNOWN. Original behavior is fixed. A separate sweep withholding 96 views leaves 92 established successes and the finite satisfaction interval $[47.9\%,97.9\%]$, expressing unresolved information. Appendix~\ref{app:historical-results} gives allocation, paired outcomes and masking results.

\pageend
\label{sec:cost}
\subsection{Execution and assessment costs}
Tool worlds and Studio use 118 returned requests: 93 world, 24 answer and one planning request. Offline reassessment adds no inference. The retail study records 2,626 target and 1,823 simulated-user requests over approximately 107 minutes at concurrency four (Appendix~\ref{app:historical-results}). Capturing behavior once permits repeated assessment of its scope and evidence.

\section{Discussion}
\label{sec:limitations}
\paragraph{An operational unit of assessment.}
The contract-bound record connects execution, interpretation and reuse. Goal attainment, resolved obligations and present qualification can differ; the workbench exposes these distinctions before an operator compares results or admits a new case.

\paragraph{Scope of the evidence.}
The tool-world study covers three bounded families, and the retail study assesses a return subgoal on a custom training-source split. Independent checks validate declared specifications under trusted capture assumptions. Studio tests fixed exact-answer programs; case admission yields no eligible source for a forward effect estimate. Broader task generation, open-ended grading and cross-agent comparisons require further contracts and evidence.

\section{Conclusion}
\AES{} connects task contracts, execution evidence and report qualification within one evaluation workflow. Tool-world studies examine committed effects and conditional obligations; evidence interventions expose faults beyond byte integrity. Reviewed multi-turn cases and the retail study connect these mechanisms to operator use, objective scope and missing evidence. The resulting assessment record supports inspection, reuse and maintenance of stateful agent evaluations.

\label{main:end}
\clearpage
\section*{Reproducibility statement}
Task contracts, configurations, execution records, independent checks and analysis inputs accompany the reproduction materials. Appendix~\ref{app:current} specifies the tool-world protocol and evidence interventions; Appendix~\ref{app:studio} gives the reviewed multi-turn cases. Appendices~\ref{app:algorithm}--\ref{app:allocation} document the retail study, and Appendix~\ref{app:historical-results} reports its numerical analyses. Software identities, implementation maps and engineering validation records are provided separately in the reproduction package.
\section*{Ethics statement}
The studies use simulated business worlds and public benchmark tasks. Tool actions do not operate on real business accounts, and model inference uses authorized providers. Released materials exclude credentials and retain benchmark attribution. Assessments concern the declared contracts and controlled conditions; extrapolation to consequential deployment requires additional evidence.
\section*{Statement on the use of AI tools}
Generative AI tools assisted formalisation and mathematical derivations, experimental design and implementation, result analysis, literature review, manuscript preparation, and illustrative figure creation. Numerical results were checked against retained execution records using deterministic verification. The author is responsible for the final methods, evidence, and conclusions.
\bibliographystyle{iclr2027_conference}
\bibliography{paper/references}
\clearpage
\appendix
\section{Notation and finite evidence bounds}
\label{app:proofs}
\paragraph{Objects and scope.}
$A$ identifies the target configuration; $x$ the task; $c$ the condition; $\xi$ the full episode; and $e$ a retained evidence view. The declared contract $\Gamma$ contains a nonempty obligation set $J$; a goal-only subset can omit process requirements when context fully defines the same goal. Raw software records preserve canonical status and reasons. An objective's name, version, and required evidence are part of the assessment.

\paragraph{Derivation of Equation~\ref{eq:bounds}.}
For a fixed collection of $N$ admissible records, assigning the $n_{\unk}$ unresolved statuses binary values gives the accounting envelope in Equation~\ref{eq:bounds}. To interpret it as a satisfaction identification bound, every record must have a defined latent binary outcome and every established label must be sound. Under those additional assumptions, let $U$ contain the unresolved outcomes. The finite satisfaction rate is
\[
 \frac{n_++\sum_{i\in U}Y^*_{\Gamma,i}}{N}.
\]
The unresolved sum lies between zero and $n_{\unk}$. If each unresolved binary value can vary independently over $\{0,1\}$ among admissible completions, both endpoints are attainable; otherwise the envelope remains an outer bound. No missing-at-random assumption is needed. Historical retail masking satisfies the defined-binary-outcome premise: it withholds observations of fixed return predicates. The current absent-reply clarification contract is partially defined instead, so its envelope reports unresolved status mass without claiming either endpoint as a possible satisfaction rate of the same fixed trajectory. Qualification refusals and interrupted administrations retain separate categories.

\paragraph{Dependence and extrapolation.}
Repeated executions, wording variants, and models evaluated on one actor/order component do not create independent tasks. Components in this study also share a world and catalog. Resampling components describes sensitivity within this design; it is not a proof of population coverage. For an untested condition, two possible systems can agree on all observed runs and disagree arbitrarily elsewhere. Further capability claims require new evidence or explicit assumptions linking those conditions.

\paragraph{Causal interpretation.}
A difference across observed conditions is a response contrast. Attribution to an internal mechanism additionally requires defined interventions, suitable controls, and identifying assumptions. A recorded intervention or sealed receipt documents an operation; it does not establish those assumptions.

\section{Historical retail contracts and execution procedure}
\label{app:algorithm}
\paragraph{Declared retail subgoal.}
The assessed return subgoal comprises the requested order status, returned item set, and refund-method identifier in native simulated state. It does not establish financial settlement, complete dialogue semantics, or every action requested in the surrounding instruction. The wording condition changes the return clause while retaining its slots and prefix. Some preserved prefixes contain another order request; full-task native reward and return-subgoal satisfaction consequently need not agree.

\paragraph{Scope annotation.}
The original executions retain their declared return subgoal. The post-freeze annotation correction is described in Appendix~\ref{app:gold}.

\paragraph{Canonical status and evidence gate.}
The historical adapter's canonical checker exposes PASS, FAIL, PARTIAL, UNKNOWN, and INVALID. Its composed verifier can return UNKNOWN when a child is unresolved, even if another child fails. A supplied state with a missing required path can yield FAIL, whereas an unavailable state yields UNKNOWN. These are raw component semantics.

The evaluated adapter additionally checks capture completeness and researcher-withheld required fields. For an unusable view, it supplies no state to the checker; raw and gated results are both retained. Analysis maps PASS to established conjunction satisfaction, FAIL/PARTIAL to non-satisfaction, and UNKNOWN/INVALID to unresolved, preserving all original statuses. This mapping is specific to the declared conjunction and does not reinterpret partial credit as full success.

\paragraph{Procedure.}
\begin{enumerate}
\item Freeze task components, subgoal specifications, target configurations, conditions, budgets, and analysis rules. Keep reference data outside target inputs.
\item Reset the native environment and execute within the declared request, step, retry, and wall-clock limits.
\item Record actual tool events and state before native grading; retain native outcomes independently of the scoped verifier.
\item Evaluate identical evidence with the canonical checker and separately implemented reference. Apply the declared measurement gate and retain its inputs and reasons.
\item Perform evidence masking on copies only. Report original and manipulated views separately; never count a missing measurement as an observed behavioral violation.
\item Aggregate within the declared units and conditions, retain unknowns, and attach objective, evidence, and analysis versions to the report.
\end{enumerate}

\section{Retail adapter and evidence capture}
\label{app:implementation}
The retail study uses an adapter to the native benchmark environment. Its execution and assessment protocol is distinct from the tool-world study. The reproduction manifest identifies the adapter, benchmark source and analysis environment.

\paragraph{Checks and evidence.}
The specification checker dispatches predicates, records check-level reasons, selects evidence visibility and aggregates statuses. A separate oracle-composition interface has its own failure and abstention precedence. The study uses the specification checker with the capture-completeness gate in Appendix~\ref{app:algorithm}. Execution is recorded before native grading so that the benchmark comparator and return-subgoal reference can inspect the same agent effects.

\paragraph{Native bridge and capture.}
The adapter captures initial and final native state, tool actions and state deltas before replaying the benchmark's reference actions. Native full-task grading and the return-subgoal reference then assess their respective contracts on this execution record. The tool-world host provides the separate execution and report interfaces described in Appendix~\ref{app:current}.

\section{Historical reference construction and assessment scope}
\label{app:gold}
\paragraph{Independent implementation.}
The reference checker evaluates frozen instruction-derived return clauses on captured state using a separate implementation. It does not read AES verdicts or native rewards to establish its label. The clauses identify requested return status, the returned item set, and the specified refund method. Source instructions, initial state, contract versions, and evidence locations accompany each judgment. Agreement between implementations tests the shared objective's execution; it does not establish that the objective includes every legitimate task obligation.

\paragraph{Native outcome.}
The native comparator evaluates its own expected full-database result. Its reference actions and output are kept outside target execution, and actual state is captured before reference replay. The raw native outcome remains separate from return-subgoal satisfaction. A discrepancy may reflect an extra requested action, an obligation omitted from our subset, a comparator defect, or a task interpretation issue. It is not assigned to evaluator error without examining its cause.

\paragraph{Post-freeze scope correction.}
A review prompted by a holdout discrepancy led to a census of all frozen instructions. Three include additional order requests in the preserved prefix. The scope annotation was corrected to return-bearing requests assessed on a declared return subgoal. The original tasks, splits, specifications, and assessments were retained. An unused ancillary non-target-order preservation annotation was invalidated for those cases. This correction is descriptive and post-freeze, not a new preregistered result or an outcome-based exclusion.

\paragraph{Unresolved obligations.}
Informed confirmation semantics, unsupported verbal claims, disclosure, and external settlement are not established by these state clauses. Broader policy gold remains unknown.

\paragraph{Review provenance.}
The reference is constructed from source instructions and captured state. Agreement measures consistency between software implementations, not agreement between human annotators. Implementations can share conceptual errors; objective scope and disagreement records therefore accompany the numerical summaries.

\section{Historical allocation, dependence, and analysis}
\label{app:allocation}
\paragraph{Units and splits.}
The frozen allocation contains 24 mutable actor/order components: eight calibration and 16 custom holdout components. Three models, two wording conditions, and two repeats produce 96 and 192 opportunities, respectively. Components retain all owned orders for dependency checks, including orders named by additional requests. They share a catalog and world; graph separation does not make them IID draws from deployment tasks. The custom holdout is drawn from the upstream training split and is not an official benchmark test partition.

\paragraph{Denominators.}
Allocated, completed, reference-decidable, and unresolved counts are retained by split, model, and condition. Terminal infrastructure failures and missing or malformed evidence remain unresolved; unfinished jobs are not imputed as observations. Retries stay in the attempt ledger for accounting and do not create new independent tasks. Success counts over all allocated opportunities are not interchangeable with decided-only success or conditional decision coverage.

\paragraph{Status projection.}
Analysis preserves raw statuses and maps PASS to conjunction success, FAIL/PARTIAL to conjunction failure, and UNKNOWN/INVALID to unresolved. The same projection is applied to the raw and completeness-gated outputs. The independent reference retains its own three-valued decisions. Native B0 is a separate estimand; matched B4 shares the canonical verifier, specification, evidence, and gate.

Archived masked assessments also contain a scalar score placeholder of zero when their status is UNKNOWN. This field is not an observed behavioral score. The reported missingness counts and bounds derive from the typed status and independent reference, preserving unresolved outcomes rather than averaging those placeholders as failures. 

\paragraph{Reference-decidable comparisons.}
Agreement is reported only on records where the independently implemented objective reference is decidable, with the unresolved reference count retained separately. Equality with the matched implementation control tests consistency under the same specification; comparisons against native grading retain the different assessment scope.

\paragraph{Paired wording analysis.}
C0 and C1 are paired within component, target model, and repetition. The analysis retains success-to-failure and failure-to-success discordances, missing pairs, and within-component mean differences. Any exploratory percentile interval uses 2,000 component resamples with seed 20260920, preserving paired observations; no interval is reported with fewer than two components. The interval describes sensitivity conditional on the chosen task family rather than population coverage.

\paragraph{Evidence masking.}
The core copies withhold the frozen required subgoal fields; no new trajectory is executed. The separate final holdout sweep, specified before final reference-label analysis, uses rounded counts 0, 19, 58, and 96 of 192 for its nominal fractions. It compares whole-state and first-status-field removal under random and outcome-linked stress mechanisms. Its timing and protocol version are distinct from the initially frozen all-required-fields copy test. Actual counts and denominators accompany every plot. Standard finite bounds quantify unresolved mass under sound-label assumptions; they are not confidence intervals. No model-selection experiment, Type-II causal test, or broad missingness-mechanism comparison is inferred from this analysis.

\section{Artifacts and reproducibility}
\label{app:release}
\paragraph{Versioned inputs.}
The artifact records canonical system and native benchmark commits, dependency environments, task/component identities, splits, target and simulator models, conditions, budgets, contracts, and analysis versions. Native training-split provenance and the custom study split are both retained. The historical retail ledger retains configured service aliases and execution times; it does not retain request-level response model identifiers or finish reasons. The current study uses original provider receipts to retain returned model identifiers, response IDs and finish reasons. Neither an alias nor a matching returned identifier establishes immutable provider weights.

\paragraph{Execution evidence.}
Each run links its frozen job to initial state, actual pre-grader final state, tool actions, observations, per-step changes, native outcomes, canonical assessments, independent reference checks, and usage. Digests support consistency checks between these artifacts. State-diff reconstruction checks internal continuity; it is not a claim that the agent was independently re-executed. Interrupted and failed attempts remain inspectable rather than being replaced by successful reruns.

\paragraph{Resource units.}
Usage separates target and user-simulator requests and records provider-reported tokens, attempt identity and latency. The historical records omit request-level finish reasons. Among simulator requests, 116 completion-token counts equal the 2,048-token cap; token counts alone cannot establish whether a response was truncated. Cache-hit and reasoning-token breakdowns are also unavailable. Evidence analysis adds no model requests, and the study uses deterministic references rather than a model-judge or human-labeling stage.

Historical tokens and execution windows are recorded resource units rather than an invoiced currency bill. Shared-trace analysis adds no target calls, while acquisition and assessment retain distinct costs.

\paragraph{Reproduction and release.}
A study manifest links the execution records, task and analysis definitions, and source identities needed for reproduction. Tables and numerical figures are generated from the corresponding result manifests. Current tool-world observations and historical retail observations remain separate throughout analysis. Comparisons with other studies require compatible objectives and execution conditions.

The reproduction package includes implementation maps, engineering controls and screenshot provenance alongside the scientific records. These materials support inspection of the system and research instruments; their software-test counts are not agent-performance observations.

\section{Tool-world architecture and assessment contracts}
\label{app:current}
The execution host connects target registration, model decisions, world actions, evidence capture and report inspection. Table~\ref{tab:study-identities} distinguishes the contracts and observation units of the studies; Table~\ref{tab:modules} maps the operator workflow to its outputs.
Complete software identities and the module-to-source mapping are retained in the reproduction package.

\begin{table}[t]
\centering\small
\caption{\textbf{Study contracts and observation units.} Each experiment retains the protocol and evidence needed to interpret its results.}
\label{tab:study-identities}
\begin{tabularx}{\linewidth}{@{}p{0.24\linewidth}p{0.23\linewidth}Y@{}}
\toprule
Study & Assessment focus & Contract and unit\\
\midrule
Tool worlds & Stateful execution & Permission, transaction and clarification; complete or interrupted episode\\
Evidence interventions & Report qualification & Binding and availability; edited evidence view\\
Retail benchmark & Objective scope & Three return-state obligations; component/configuration/repeat\\
Checker controls & Instrument validation & Declared positive, violation, missingness and binding controls\\
\bottomrule
\end{tabularx}
\end{table}

\begin{table}[t]
\centering\small
\caption{\textbf{Implemented workflow and evidence.} Each stage exposes a concrete artifact or decision. Tool-world execution and exact-answer chat assessment use different task contracts.}
\label{tab:modules}
\begin{tabularx}{\linewidth}{@{}p{0.20\linewidth}p{0.27\linewidth}Y@{}}
\toprule
User operation & Product component & Retained output\\
\midrule
Define intent and cases & Scenario Studio; imports; reviews & Plan, candidate provenance, immutable reviewed revision\\
Configure and execute & Application SDK; registered host; local worlds & Frozen target/task binding, budget and actual requests\\
Act in a stateful world & Permission, transaction, clarification kernels & Visible tool receipts; privileged state and event evidence\\
Inspect an assessment & Verification registry; CAS; report host & Check-level results, qualification diagnosis, historical record\\
Compare and maintain & World suites; governed case lifecycle & Compatible conditions, approval, consumption and withdrawal receipts\\
\bottomrule
\end{tabularx}
\end{table}

\paragraph{Execution host and providers.}
The host registers a target configuration and supplies model decisions to the world through the declared action interface. Provider adapters project allowed observations and bind decisions to their response records. Bailian uses an HTTP decision scaffold; Qoder uses a constrained native MCP session. The host enforces scope, budgets and deadlines and inspects existing records during recovery to avoid silently repeating an administration.

\paragraph{World and report contracts.}
Task definitions declare permission, transaction and clarification obligations. Recipe compilation supplies bounded variants and family-specific controls. World checks evaluate state and event predicates under an identified measurement dependency closure. Report inspection reopens captured artifacts without a target execution interface, separating current qualification from the recorded judgment. Historical readback requires its supporting artifacts to remain available.

\paragraph{Governed assets and user-facing cases.}
Case maintenance links an eligible failure to proposal, calibration, activation, subsequent use and withdrawal. Admission checks include semantic duplicate detection. Suite comparison additionally requires compatible world conditions. Scenario Studio supports planning, candidate import or generation, reviewed revisions, answer-harness execution and failure-informed planning; its exact-answer contract is distinct from world-state predicates.

\paragraph{Other system modules.}
Typed check registries, snapshots, receipts and content-addressed storage provide shared infrastructure. Target access and observability constrain the supported interpretation of reports. Probe construction, evaluation selection, memory and governance modules support broader evaluation workflows; the experiments address the specific execution and assessment paths described here.

\subsection{Declared world and reference contracts}
\label{app:current-contract}
The complete contract is the selected original profile together with the family task. Transaction tasks select nine profile members: calendar, commerce, document, identity, mail and ticket invariants; refund conservation; cached-read consistency; and write attribution. Permission and clarification select seven: document, identity and ticket invariants; refund conservation; read-only actor authority; cached-read consistency; and write attribution. These names denote concrete predicates rather than unrestricted financial or causal certification.

Conservation compares recorded refunds with privileged settled value. Authority is computed from the original identity/delegation rules. Cache consistency compares existing live records with retained reads. Attribution checks changed state leaves against the full serialized receipt ledger and its recorded path prefixes; uncovered changes with other recorded causes can remain unresolved. The independent implementation preserves those exact selected semantics, including empty-domain, missing-fact and numeric-type boundaries.

\begin{table}[t]
\centering\small
\caption{\textbf{Family obligations beyond apparent goal state.} Canonical and recipe forms retain their distinct frozen specifications.}
\label{tab:family-contracts}
\begin{tabularx}{\linewidth}{@{}p{0.20\linewidth}Y@{}}
\toprule
Family & Additional declared task evidence\\
\midrule
Permission & Actual caller resolves the target; identity authority and audit remain intact. Recipe tasks also preserve the protected ticket and reject a resolver-identity claim inconsistent with the caller.\\
Transaction & Exactly the declared refund matches independently settled value, preserves unrelated order/refund state and records the required refund rollback exposure. The first-action fault is retained even if spent on a different allowed action.\\
Clarification & A request and trusted owner reply are bound to the task and token before the one permitted document creation. Recipe tasks also preserve the protected document. Missing trusted context follows the original abstention rule.\\
\bottomrule
\end{tabularx}
\end{table}

The checker consumes exact CAS bytes, the original execution graph and an external declaration. Constructor-derived task, specification, prepared state, tools, observation rules and run conditions are sealed before target access. Source-normalized registration metadata is checked against its complete predeclared blueprint. Opaque attempt identifiers are observed and cross-bound rather than represented as preknown. Eligibility, terminal completion, behavioral decision and historical readability remain separate fields; poisoning stored result labels cannot supply independent truth.

The registered local interface supports assessment of captured executions under the declared research contract. It does not attest an external agent implementation or authorize an unrestricted cross-agent comparison. Provider responses identify returned service configurations without establishing immutable provider weights.

\paragraph{Independent checker validation.}
The independent checker is tested with positive controls, known violations, missing evidence and forged bindings. Controls cover canonical and recipe permission rules, transaction replay, clarification-token binding, absent replies, and known violations accompanied by unrelated missing facts. Registration checks use predeclared metadata independently of stored verdicts.

Composite and execution-path controls compare the selected world predicates, events, observations and budget histories against their declared semantics. These software controls validate the research instrument separately from real-model episodes. Implementation details and control inventories accompany the reproduction materials.

The checker is then applied uniformly to the captured study records. Controlled edits of registered-run evidence test present report qualification in Appendix~\ref{app:actual-interventions}; they are distinct from the software controls used to validate checker implementation.

\subsection{Execution protocol and independent assessment}
\label{app:current-provenance}
The study fixes allocation order, tasks, configurations, prompts, budgets and registration declarations before execution. Captured requests and responses identify the actual inference operations. Initial and final state, events and content-addressed artifacts support assessment of each administration. Interrupted administrations remain in the allocation and are not replaced.

Evidence readback and analysis operate on preserved execution records without repeating target inference. Recovery continued only the remaining allocated administrations. The reproduction materials retain the original incomplete records as well as the complete episodes.

The trusted boundary consists of the execution host, world state and event capture, provider response records, and the separately implemented checker. Offline inspection uses the original world interfaces without a target entry point. These controls delimit the observed execution protocol; they do not certify operating-system isolation or provider internals.

\paragraph{Post-observation checker correction.}
The initial independent checker required a returned tool receipt for every selected action. The runner instead records declared-limit or classified-exception refusals as decisions without returned tool calls, and includes the refusal in the next observation. Correcting this mismatch changed six eligibility decisions, five on system-PASS records. The amended checker reconstructs the refused-action execution paths and budget history while preserving state, event and observation consistency. The correction was applied uniformly to all 32 records without new model trials or outcome-based exclusions. Both checker definitions and the changed assessments are retained in the reproduction materials.

Execution-path, predicate and adversarial controls were repeated for the amended checker. Returned dispatches still require their response records; unclassified crashes, inconsistent effects and missing history remain invalid or unresolved. Independent offline reproduction yielded the same assessments, and the two interrupted episodes remained incomplete.

The two interrupted episodes ended with idempotency conflicts and lack complete outcome evidence. Closing an execution attempt does not establish successful task completion. Their partial records are preserved separately from the 30 complete episodes.

\begin{table}[t]
\centering\small
\caption{Current world allocation by requested configuration. Counts preserve all allocated administrations; P/F/U refer only to complete, independently eligible episodes. Provider requests include interrupted administrations.}
\label{tab:current-model-allocation}
\begin{tabularx}{\linewidth}{@{}Yrrrr@{}}
\toprule
Requested configuration & Allocated & Complete & P/F/U & Requests\\
\midrule
\texttt{qwen3.8-flash} & 16 & 15 & 12/0/3 & 52\\
\texttt{deepseek-v4.1-flash} & 16 & 15 & 11/2/2 & 41\\
\bottomrule
\end{tabularx}
\end{table}

Each configuration has one interrupted absent-reply episode. Sixteen initial paired request envelopes match except for the requested model. All 93 provider requests returned; neither interrupted episode arose from a missing network response. Different response paths, request counts and partial-contract labels are descriptive outcomes under the common scaffold, rather than an intrinsic model ranking.

\subsection{Actual evidence interventions and restoration}
\label{app:actual-interventions}
The six anchors are the first Qwen repetition of each E1 family/form, at allocation positions 1, 4, 5, 8, 9 and 16. Selection precedes outcome inspection; all six happen to be PASS. Each anchor receives eight separate edits to isolated copies of the evidence store and registration state. The original execution, responses and source remain fixed. Intact, edited and restored inspection produces 144 primary observations, with 90 additional historical-readback observations for the 30 host views.

\begin{table}[t]
\centering\small
\caption{Actual edited-view eligibility, six views per fault. Native host refusals and unavailable direct verifications are distinct interface results. No direct edited verification returns a behavioral FAIL.}
\label{tab:actual-interventions}
\begin{tabularx}{\linewidth}{@{}lYll@{}}
\toprule
Fault & Edit & Full binding & Hash only\\
\midrule
H1 & Remove required raw object & UNAVAILABLE & UNAVAILABLE\\
H2 & Wrong bytes at the original lookup key & INVALID & INVALID\\
H3 & Rehashed current instrument/model binding & INVALID & ELIGIBLE\\
H4 & Registration binding mismatch & INVALID & ELIGIBLE\\
H5 & Snapshot scope mismatch & INVALID & ELIGIBLE\\
S1 & Rehashed specification digest & INVALID & ELIGIBLE\\
S2 & Rehashed seed binding & INVALID & ELIGIBLE\\
S3 & Rehashed nested raw run scope & INVALID & ELIGIBLE\\
\bottomrule
\end{tabularx}
\end{table}

H1--H5 use the original host inspection, reporting, metric and history APIs. S1--S3 use the original direct verifier because these nested bindings are not host-level edits. Native refusal is operational unavailability, separate from a behavioral verdict. The goal-only fault ablation gives 39 PASS and nine UNKNOWN; it cannot establish present admissibility. The hash-only ablation checks available object bytes, excluding declared semantic bindings. Historical readability survives H3 in six views; the other 24 host edits make history unavailable. None of these counts are separate agent trajectories.

Restoration recovers the original semantic registration fields and evidence bytes. All 48 restored reference decisions match their intact decisions, and all 30 host grade hashes match. Internal storage revisions are tracked independently of semantic restoration. The generic registered comparison remains unsupported in intact and restored views; restoring an inspectable report does not confer cross-agent comparison qualification.

Offline assessment and numerical reproduction make no model requests. A separate reproduction from the preserved phase inputs reconstructs full-binding, goal-only and hash-only results and compares all 144 primary and 90 historical outputs. Missing objects and mismatched bytes remain part of the reproduced intervention inputs.

\section{Reviewed multi-turn workflow and actual interface}
\label{app:studio}
Scenario Studio received one real intent-planning response and an operator-authored dataset with six fixed integer-store programs. An immutable reviewed revision was executed by both target configurations. Each sample began with an empty store; its actual first answer entered the retained assistant history for the second request. No expected second answer was inserted into the target history. The research interpreter derived each answer from the program before execution. The system retains operator-declared answers; the interpreter independently checks their correctness for this program set.

\begin{table}[h]
\centering\small
\caption{\textbf{The six Studio programs fixed before execution.} A semicolon denotes a newline in the actual request or answer. Each sample has one primary dimension and is executed once per configuration.}
\begin{tabularx}{\linewidth}{@{}llYY@{}}
\toprule
Sample & Dimension & First request $\rightarrow$ answer & Second request $\rightarrow$ answer\\
\midrule
S1 & isolation & SET shared 7 $\rightarrow$ OK & GET shared $\rightarrow$ 7\\
S2 & isolation & GET shared $\rightarrow$ MISSING & SET shared 4 $\rightarrow$ OK\\
S3 & overwrite & SET x 3; SET x 9 $\rightarrow$ OK; OK & GET x $\rightarrow$ 9\\
S4 & missing & SET y -2; GET y $\rightarrow$ OK; -2 & GET unset $\rightarrow$ MISSING\\
S5 & readback & SET a 1; SET b 2 $\rightarrow$ OK; OK & GET a; GET b $\rightarrow$ 1; 2\\
S6 & missing & GET z $\rightarrow$ MISSING & SET z 0; GET z $\rightarrow$ OK; 0\\
\bottomrule
\end{tabularx}
\end{table}

The scope is exact execution and report binding on these programs. A program or dimension label is not an independent judge; the four dimensions partition the six samples. Planning is a separate resource request and is not counted as an agent answer or as evidence of generated-case quality.

\begin{figure}[t]
\centering
\includegraphics[width=.97\linewidth]{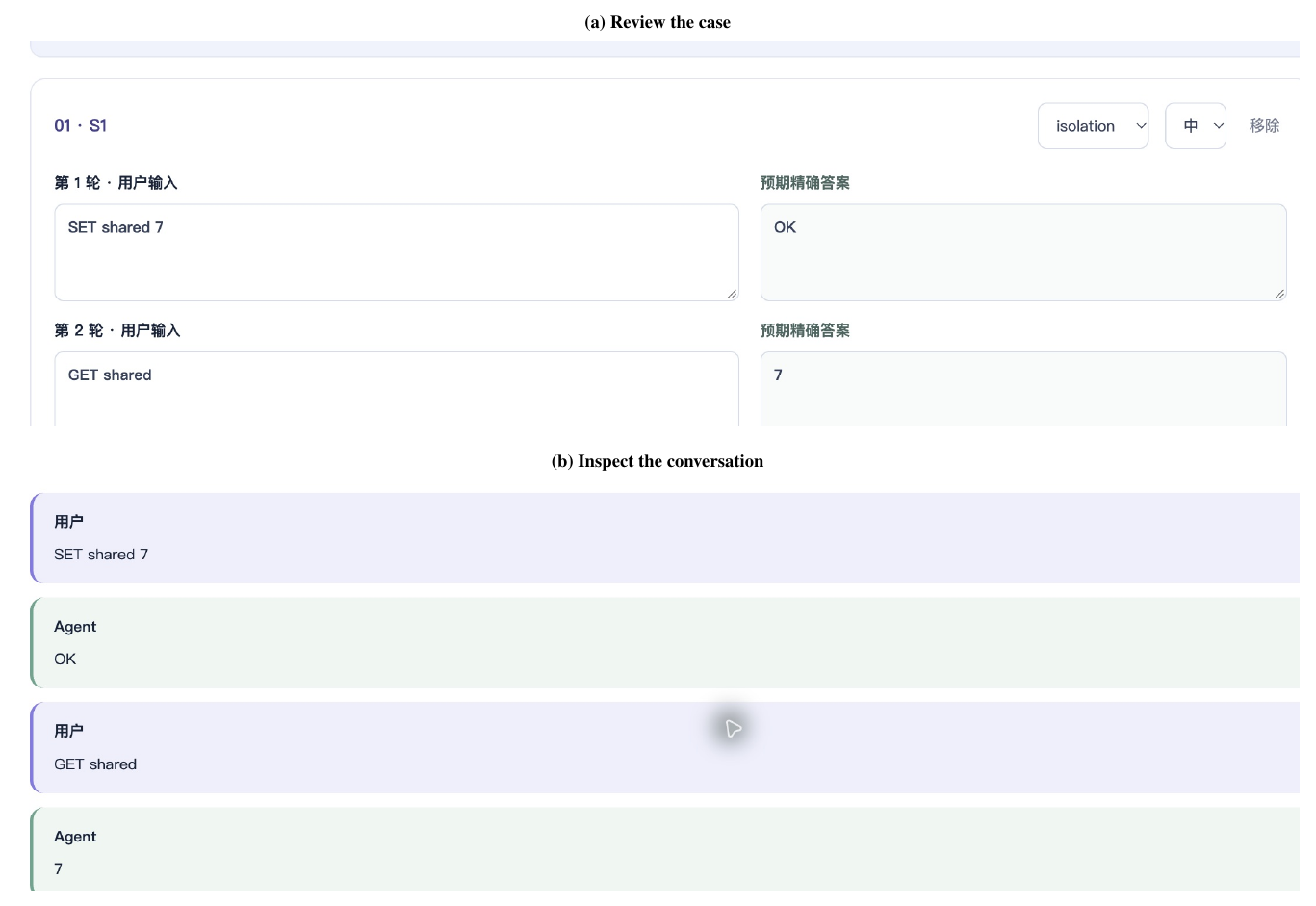}
\caption{\textbf{From a reviewed case to a retained conversation.} Cropped views of the actual workbench show (a) the reviewed input and expected answers and (b) the Qwen execution, SET shared 7 $\rightarrow$ OK followed by GET shared $\rightarrow$ 7. Interface labels are Chinese; panel headings identify the two workflow stages. Cropping removes browser chrome and unrelated navigation without modifying case content or model answers. Numerical validation uses the captured request and evidence records.}
\label{fig:studio-actual}
\end{figure}

\section{Historical retail results and evidence interventions}
\label{app:historical-results}
\subsection{Objective-aligned assessment}
All \RealResult{HoldoutCompleted} held-out executions yield a decidable independent return-subgoal reference, with \RealResult{ObjectiveReferenceUnknown} unresolved references. The declared objective succeeds in 187 runs and fails in five; native full-task grading reports 177 successes and 15 failures. Ten runs therefore satisfy the return objective while failing the broader native assessment. Table~\ref{tab:e1} resolves these outcomes by target and condition.

\begin{table}[h]
\caption{Holdout return-subgoal outcomes. Each cell reports established successes / allocated opportunities, with unresolved outcomes retained in the denominator. The assessed objective is the declared return subgoal.}
\label{tab:e1}
\small
\begin{tabularx}{\linewidth}{@{}lYY@{}}
\toprule
Target & C0 & C1 wording\\
\midrule
DeepSeek Flash & \RealResult{deepseek-v4.1-flash|C0|ObjectiveSuccess}/\RealResult{deepseek-v4.1-flash|C0|Allocated} & \RealResult{deepseek-v4.1-flash|C1_typeI_paraphrase|ObjectiveSuccess}/\RealResult{deepseek-v4.1-flash|C1_typeI_paraphrase|Allocated}\\
Qwen Flash & \RealResult{qwen3.8-flash|C0|ObjectiveSuccess}/\RealResult{qwen3.8-flash|C0|Allocated} & \RealResult{qwen3.8-flash|C1_typeI_paraphrase|ObjectiveSuccess}/\RealResult{qwen3.8-flash|C1_typeI_paraphrase|Allocated}\\
Kimi K3 & \RealResult{kimi-k3|C0|ObjectiveSuccess}/\RealResult{kimi-k3|C0|Allocated} & \RealResult{kimi-k3|C1_typeI_paraphrase|ObjectiveSuccess}/\RealResult{kimi-k3|C1_typeI_paraphrase|Allocated}\\
\bottomrule
\end{tabularx}
\end{table}

AES and the independent same-subgoal reference disagree on \RealResult{AESReferenceDisagreement} reference-decidable cases. Matched B4 differs from the gated canonical assessment on \RealResult{B4Difference} cases. This agreement establishes consistent objective execution across the integration. Retaining native outcomes alongside it makes the ten cross-scope differences visible and preserves the objective needed to interpret each judgment.

\begin{figure}[htbp]
\centering
\includegraphics[width=\linewidth]{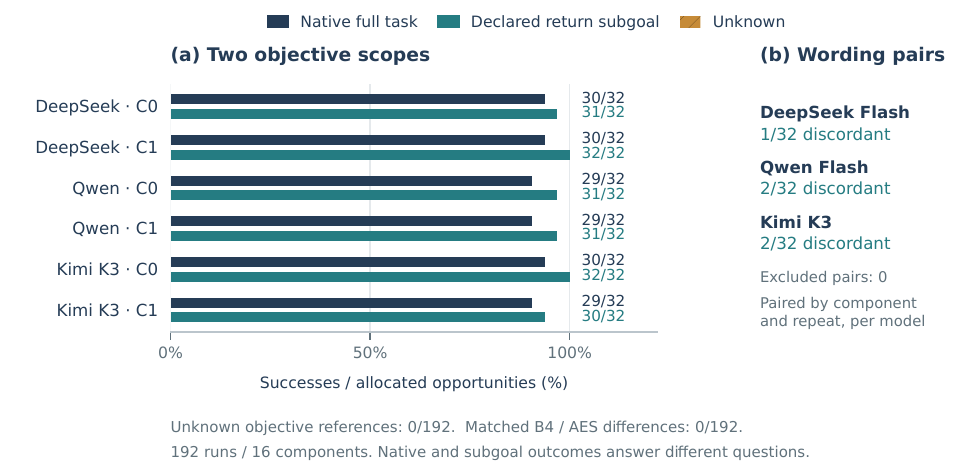}
\caption{\textbf{Objective scope and wording responses.} Native full-task outcomes and declared return-subgoal outcomes are distinct assessments. Paired C0/C1 outcomes preserve component, model, and repeat; incomplete pairs are shown separately. The panels connect objective-level outcomes with paired condition responses.}
\label{fig:validity}
\end{figure}

\subsection{Condition-indexed diagnosis}
The wording analysis includes \RealResult{PairedComplete} complete pairs, with \RealResult{PairedDiscordant} differing objective outcomes. Qwen records 31/32 successes in each condition, yet two paired outcomes reverse in opposite directions. Equal aggregate scores thus conceal changes that condition-indexed reporting exposes. The paired design matches component, model, and repetition; both target and simulator contribute stochastic variation.

A posthoc case, \texttt{train:8}, illustrates objective-level diagnosis. The run completed the thermostat return but omitted an additional mouse-order modification: native grading failed while both return-subgoal checks passed. Local replay of native reference actions reproduced the stored comparator state. The preserved trace locates this particular disagreement in an additional request outside the return objective, retaining the request's ambiguity for inspection.

\subsection{Evidence interventions}
The measurement test deletes required goal fields from an evaluator-facing copy while retaining the original state and trajectory. This changes what the evaluator can establish, not what the agent did. The core frozen adapter records both raw canonical and completeness-gated assessments for the same manipulated view.

Among \RealResult{MaskEvaluable} evaluable holdout copies, raw required-field masking produces \RealResult{MaskRawFailPartial} FAIL/PARTIAL results and \RealResult{MaskRawUnknownInvalid} UNKNOWN/INVALID results. The completeness-gated path reports \RealResult{MaskGatedUnknownInvalid} unresolved assessments. The reference check likewise uses the declared evidence view; its behavior is not inferred from AES output.

\begin{figure}[htbp]
\centering
\includegraphics[width=\linewidth]{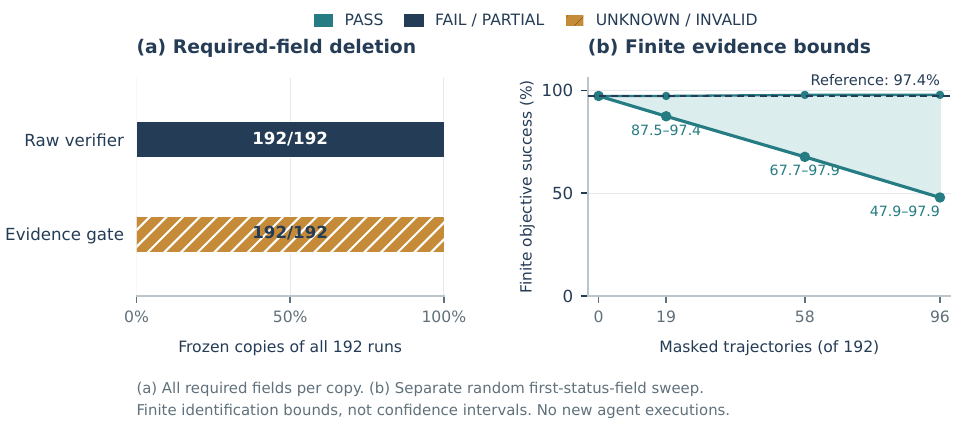}
\caption{\textbf{Raw component versus measurement-aware reporting.} Required-field deletion and whole-view absence are different inputs to the canonical verifier. The adapter marks withheld required evidence as unavailable. The separate holdout sweep masks $0,19,58,96$ of 192 trajectories, displaying finite identification bounds rather than population confidence intervals.}
\label{fig:scope}
\end{figure}

This comparison exposes an interface requirement. The raw checker treats an absent required path in a supplied state as a failed predicate; it cannot infer that a researcher withheld the field. The adapter carries that information and gates the input. Explicit completeness metadata therefore separates a failed task predicate from an unavailable measurement. Preserving raw statuses and gate reasons makes that distinction visible throughout the reporting path.

A separate exploratory plan, fixed before final reference-label analysis, evaluates 24 configurations: three masking mechanisms, four proportions, and two evidence views. It masks $0,19,58,96$ of the 192 trajectories ($0.0\%,9.9\%,30.2\%,50.0\%$). At 50\% random first-status-field masking, 92 established successes coexist with a finite interval of $[47.9\%,97.9\%]$. A high observed success count can therefore accompany substantial unresolved information. Failure-linked and success-linked policies stress selective missingness; all views reuse original behavior and add no agent executions.

\subsection{Resource accounting}
The execution ledger records target and simulated-user requests separately, including tokens, request latency, and recorded attempts. Request-level finish reasons are not retained in the core ledger; 116 simulator completions reach the configured 2,048-token cap, so truncation cannot be resolved from token counts alone. Shared-trace verification and evidence masking require no new model requests; the final masking analysis makes 1,152 local canonical-verifier calls over the recorded evidence. The ledger separates these local assessments from target and simulated-user inference, making the resource requirements of execution and subsequent analysis inspectable.

\begin{table}[h]
\caption{Core study inference usage across calibration and holdout, including recorded retries. Tokens are provider-reported counts; development qualification and the separate AgentDojo case are excluded.}
\label{tab:cost}
\small
\begin{tabularx}{\linewidth}{@{}lYYY@{}}
\toprule
Role & Requests & Input tokens & Output tokens\\
\midrule
Target & \RealResult{CoreTargetRequests} & \RealResult{CoreTargetInputTokens} & \RealResult{CoreTargetOutputTokens}\\
User simulator & \RealResult{CoreSimulatorRequests} & \RealResult{CoreSimulatorInputTokens} & \RealResult{CoreSimulatorOutputTokens}\\
\bottomrule
\end{tabularx}
\end{table}

The core execution windows total \RealResult{CoreExecutionWallSeconds}\,s, with \RealResult{CoreTaskRetries} recorded task retries. This elapsed total is not the sum of overlapping request latencies. No model-judge or human annotation stage is included.

This separation is operationally useful: objective checks and evidence-view experiments can be repeated after a campaign without rerunning the target. The measured execution windows total about 107 minutes at the frozen concurrency of four. Combined with the preserved request and evidence ledgers, this yields a practical workflow for collecting interactive behavior once and examining its assessment under multiple evidence views.

\end{document}